\documentclass[letterpaper, 10 pt, conference]{ieeeconf}  

\IEEEoverridecommandlockouts                                                                                        
\usepackage{amsmath} \usepackage{amssymb}  \usepackage{array, multirow}
\usepackage[pdftex]{graphicx}
\graphicspath{{./figures/}}
  \DeclareGraphicsExtensions{.pdf,.jpg,.jpeg,.png}
	
\usepackage[caption=false,font=footnotesize]{subfig}
	
\usepackage{url}

\usepackage{hyperref}
\usepackage[nameinlink,capitalize]{cleveref}

\usepackage{tikz,xcolor,hyperref}
\definecolor{lime}{HTML}{A6CE39}
\DeclareRobustCommand{\orcidicon}{
	\begin{tikzpicture}
	\draw[lime, fill=lime] (0,0) 
	circle [radius=0.16] 
	node[white] {{\fontfamily{qag}\selectfont \tiny ID}};
	\draw[white, fill=white] (-0.0625,0.095) 
	circle [radius=0.007];
	\end{tikzpicture}
	\hspace{-2mm}
}
\foreach \x in {A, ..., Z}{\expandafter\xdef\csname orcid\x\endcsname{\noexpand\href{https://orcid.org/\csname orcidauthor\x\endcsname}
			{\noexpand\orcidicon}}
}

\title{\LARGE \bf
Transfer Importance Sampling – How Testing Automated Vehicles in Multiple Test Setups Helps With the Bias-Variance Tradeoff*
}

\author{Max Winkelmann$^{1}$\orcidA{}, Constantin Vasconi$^{1}$, and Steffen Müller$^{2}$\thanks{*This work was supported in part by IAV GmbH, 10587 Berlin, Germany.}\thanks{$^{1}$Max Winkelmann and Constantin Vasconi are with IAV GmbH
        {\tt\small \{forename.surname\}@iav.de}}\thanks{$^{2}$Steffen Müller is with Department of Automotive Engineering, Technische Universität Berlin, 13355 Berlin, Germany
        {\tt\small steffen.mueller@tu-berlin.de}}}

\newcommand{\x}{\boldsymbol{x}}
\newcommand{\event}{\mathcal{E}}
\newcommand{\model}{\mathcal{M}}
\newcommand{\metamodel}{\widetilde{\mathcal{M}}}
\newcommand{\deriveq}{q}
\newcommand{\estimateell}{\ell}
\newcommand{\transfer}{\mathcal{T}}

\begin{document}

\maketitle
\thispagestyle{empty}
\pagestyle{empty}
\setcounter{footnote}{2} 

		\begin{abstract}
The promise of increased road safety is a key motivator for the development of automated vehicles (AV). Yet, demonstrating that an AV is as safe as, or even safer than, a human-driven vehicle has proven to be challenging. Should an AV be examined purely virtually, allowing large numbers of fully controllable tests? Or should it be tested under real environmental conditions on a proving ground? Since different test setups have different strengths and weaknesses, it is still an open question how virtual and real tests should be combined. On the way to answer this question, this paper proposes transfer importance sampling (TIS), a risk estimation method linking different test setups. Fusing the concepts of transfer learning and importance sampling, TIS uses a scalable, cost-effective test setup to comprehensively explore an AV's behavior. The insights gained then allow parameterizing tests in a more trustworthy test setup accurately reflecting risks. We show that when using a trustworthy test setup alone is prohibitively expensive, linking it to a scalable test setup can increase efficiency – without sacrificing the result's validity. Thus, the test setups' individual deficiencies are compensated for by their systematic linkage.
\end{abstract}

\section{Introduction}
It is both a promise of the developers and a requirement of society that automated vehicles (AV) do not negatively affect road safety.
In this context, ISO 21448~\cite{iso_central_secretary_road_2019} requires that explicit acceptance criteria (AC) be defined regarding the risks posed by an AV's actions.
In line with society's mandate to protect human life, a popular AC is that an AV may not cause more fatalities than a human-driven vehicle.
However, fulfilling the AC requires estimating (an upper bound of) the fatality rate of an AV, which is quite challenging.

Since accidents are rare, estimating a fatality rate using naturalistic-field operational tests (N-FOT), where an AV replicates human drivers' routes in real-world traffic, would require billions of kilometers to reach statistical significance~\cite{kalra_driving_2016}.
Thereby, the deficiencies of N-FOTs are versatile: tests can not be controlled and are real-time, other road users would be endangered, and parallelization requires additional vehicles and safety drivers.
Hence, scenario-based testing~\cite{riedmaier_survey_2020} aims to accelerate risk estimation by focusing on hazardous scenarios such as jaywalking, which can be examined in different test setups~\cite{huang_autonomous_2016}. Each test setup can overcome various disadvantages of N-FOTs~\cite{steimle_toward_2022}.
For example, 3D simulations are well suited for investigating the influences of weather conditions at a reasonable cost. Conversely, highly dynamic maneuvers can be better investigated on proving grounds.

The test setups' deficiencies influence the validity of test results and hence pose a key challenge when gathering evidence for an AV's safety~\cite{huang_synthesis_2018}. Therefore, great effort is made to mitigate test setups' deficiencies, e.g., by creating more realistic simulations. Nevertheless, it can be assumed that there will always be scalable but abstract (e.g., purely virtual) test setups on the one hand and trustworthy but costly (e.g., proving ground) test setups on the other hand.
We, therefore, compensate for the deficiencies of different test setups by linking them using an approach fusing the concepts of transfer learning (TL) and importance sampling (IS).

In this paper, we introduce transfer importance sampling (TIS), a method for efficient risk estimation. In our demonstration, we derive an understanding of an AV's behavior from a scalable 2D simulation and use this understanding to evaluate risks in a more trustworthy 3D simulation.
The 3D simulation ensures the validity of the results, while the insights gained from the 2D simulation increase efficiency.
Specifically, we apply TIS to a jaywalking scenario with varying weather conditions, where the 2D simulation can not model all influences examined in the 3D simulation.
As such, the method is also applicable for linking other test setups, for example, 3D simulations to proving ground tests.

\section{Related Work}
\begin{figure}[!t]
	\centering
	\includegraphics[width=3.4in]{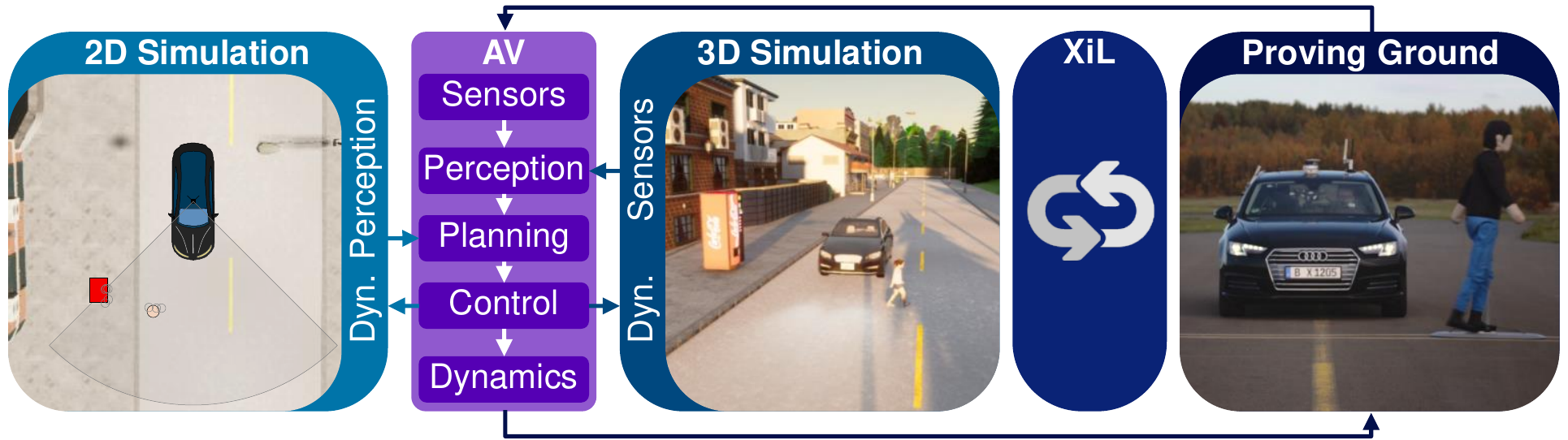}
	\caption{Examples of closed-loop test setups for AV. While for proving ground tests, the AV, including all of its components, can be tested, most test setups use models of the environment and AV's components. A typical 3D simulation uses models of the environment, sensors, and dynamics. A 2D simulation further represents the AV's perception components as a model. Purely virtual tests (without the AV's hardware) can easily be parallelized.}
	\label{fig:test_setups}
\end{figure}

When estimating risks, the choice of test setup is of great importance.
With reference to N-FOTs, a test setup is a model $\model$ allowing closed-loop testing of an AV (see \cref{fig:test_setups}).
The proportion of modeled components and the quality of the models influence a risk estimate's bias (the systematic error between the risk estimated in tests and the risk observed after release).
The scalability of a test setup influences the number of possible test runs and thus the variance (the confidence interval's width) of a risk estimate. However, the variance also depends on how test runs are parameterized.

In scenario-based testing, test runs are parameterized by a vector $\x$ containing parameters such as distances and weather conditions. Accordingly, $p(\x)$ describes the likelihood of parameterizations expected in the AV's operational domain.

\subsection{Monte Carlo Sampling}
For Monte Carlo (MC) sampling, $N_\ell$ test runs are parameterized with samples from $p(\x)$.
The expected likelihood of occurrence $\ell_{\event \model}$ of the event $\event$ (e.g., a collision) based on the model $\model$ is calculated using \cref{eq:mc_mean}. The indicator function $J$ returns 1 if $\event$ occurs and 0 otherwise.
\begin{equation} \label{eq:mc_mean}
	\ell_{\event \model} = \frac{1}{N_\ell} \sum_{i=1}^{N_\ell} J(\event | \model, \x_i)
\end{equation}
Here, \cref{eq:mc_variance} shows that the relative standard deviation $\sigma_{\ell\ \mathrm{rel}}$ increases with decreasing $\ell_{\event \model}$ and reduces with $\sqrt{1/N_\ell}$, indicating that estimating small $\ell_{\event \model}$ requires many test runs.
\begin{equation} \label{eq:mc_variance}
	\sigma_{\ell\ \mathrm{rel}} = \sqrt{\frac{\mathrm{var}\{ \ell_{\event \model} \}}{\ell_{\event \model}^2}} = \sqrt{\frac{1 - \ell_{\event \model}} {\ell_{\event \model}\ N_\ell}}
\end{equation}
Accordingly, \cite{yang_development_2010} simulates more than 14 billion kilometers and \cite{woodrooffe_evaluation_2014} 1.5 million test runs to assess collision warning systems. In both cases, generating such amounts of data was made possible by using very abstract, equation-based models.

\subsection{Importance Sampling}\label{subsec:IS}
To efficiently estimate risks even in the case of small $\ell_{\event \model}$, IS does not sample from $p(\x)$ but from a proposal density $q(\x)$, under which $\event$ occurs more frequently.
The observations are then scaled to $p(\x)$ using \cref{eq:is_mean}.
\begin{equation} \label{eq:is_mean}
	\ell_{\event \model} = \frac{1}{N_\ell} \sum_{i=1}^{N_\ell} J(\event | \model, \x_i)\ \frac{p(\x_i)}{q(\x_i)}
\end{equation}
The estimate is unbiased if $q(\x)>0\ \forall\ p(\x)\ J(\event | \model, \x)>0$.\\Using the optimal proposal density $q^*(\x)$ in \cref{eq:q_optimal}, an exact risk estimate could be achieved with a single test run.
\begin{equation} \label{eq:q_optimal}
	q^*(\x) = \frac{P(\event | \model, \x)\ p(\x)}{\int P(\event | \model, \x)\ p(\x)\ d\x} = \frac{P(\event | \model, \x)\ p(\x)}{\ell_{\event \model}}
\end{equation}
However, since in $q^*(\x)$, $\ell_{\event \model}$ appears (the very quantity that is to be determined), $q^*(\x)$ is unknown.
Nevertheless, a sufficiently good $q(\x)$ can often be defined by experts.
This can involve manipulating $p(\x)$ through shifting~\cite{gietelink_probabilistic_2005} or the exclusion of frequently occurring parameterizations~\cite{zhao_accelerated_2016}.
Alternatively, the initial parameters of test runs can be used to anticipate the risk~\cite{akagi_risk-index_2019} and exclude uncritical test runs~\cite{wang_combining_2020}.

For complex scenarios, however, experts may not be able to anticipate which test runs are critical or come up with test runs critical for humans, but not the actual AV under test.

\subsection{Adaptive Importance Sampling}
By linking $q(\x)$ to the scenario and AV under test, adaptive importance sampling (AIS) can be applied to complex scenarios and often follows the cross-entropy (CE) method~\cite{rubinstein_cross-entropy_1999}:
First, a family of distributions for $q(\x)$ is chosen, which can be parametric, mixture, or non-parametric depending on the desired flexibility. Next, test runs are performed to derive $q(\x)$'s parameters. Finally, the actual risk estimation is done.

\cref{tab:ais} shows that AIS can be applied to complex scenarios. A direct comparison of the approaches is difficult due to differences in scenarios, risk levels, and termination criteria. Yet, the test setups are either very abstract or parallelized, most likely due to the high numbers of test runs.

However, since the number of test runs $N_q$ spent to derive $q(\x)$ is mostly larger than the number of test runs $N_\ell$ spent for the actual risk estimation, there is room for improvement:
The derivation of $q(\x)$ and the actual estimation of $\ell_{\event \model}$ can involve two models, $\model_\deriveq$ and $\model_\estimateell$, instead of one model $\model$. To ensure low bias, only $\model_\estimateell$ must accurately reflect risks.

\begin{table}[!t]
		\renewcommand{\arraystretch}{1.18}
	\setlength{\tabcolsep}{4pt}
	\begin{center}
		\caption{Adaptive Importance Sampling Approaches}
		\label{tab:ais}
		\begin{tabular}{|l|l|l|c|c|c|}
			\hline
			$q(\x)$                     & \textbf{Scenario} & \textbf{Test Setup} & $N_q$ & $N_\ell$ & \textbf{Src.}                    \\
			\hline 			\multirow{2}{*}{parametric} & Lane-Change       & equation-based      & 30k   & 12k      & \cite{zhao_accelerated_2017}     \\ 									\cline{2-6}
						\cline{2-6}
			                            & Multi-Vehicle     & 3D (distributed)    & 20k   & 50k      & \cite{okelly_scalable_2018}      \\

			\hline 			\multirow{2}{*}{mixture}    & Lane-Change       & equation-based      & 24k   & 7840     & \cite{huang_accelerated_2018}    \\
			\cline{2-6}
						                            & Multi-Vehicle     & equation-based      & 40k   & 2928     & \cite{jesenski_creation_2021}    \\

			\hline 			\multirow{2}{*}{non-}       & Car-Following     & equation-based      & 4782  & 547      & \cite{gietelink_adaptive_2006}   \\
			\cline{2-6}
			\multirow{2}{*}{parametric} & Emergency Brake   & equation-based      & 10k   & 10k      & \cite{de_gelder_assessment_2017} \\
			\cline{2-6}
			                            & Multi-Vehicle     & 3D (cloud)          & n/a   & n/a      & \cite{norden_efficient_2020}     \\
									\hline
		\end{tabular}
	\end{center}
\end{table}

\section{Transfer Importance Sampling}
In the following, we present the steps that TIS uses to distribute risk estimation across two test setups, thereby compensating for the deficiencies of the individual test setups.

\subsection{Metamodel-Based Importance Sampling}\label{subsec:mmis}
To derive the process for TIS, we first introduce metamodel-based IS~\cite{dubourg_metamodel-based_2013}, a form of AIS. Based on $q^*(\x)$ in \cref{eq:q_optimal}, $\model$ is substituted by its metamodel $\metamodel$.
\begin{equation} \label{eq:q_mm}
	q_{\metamodel}(\x) = \frac{P(\event | \metamodel, \x)\ p(\x)}{\int P(\event | \metamodel, \x)\ p(\x)\ d\x} = \frac{P(\event | \metamodel, \x)\ p(\x)}{\ell_{\event \metamodel}}
\end{equation}
While $q_{\metamodel}(\x)$ is not optimal, metamodels' high computational efficiency allows estimating the integral in \cref{eq:q_mm} using MC sampling.
Here, $\metamodel$ may be a classification model which predicts the event's (e.g., a collision's) probability after being trained with a set of test results; to ensure an unbiased estimate (see \cref{subsec:IS}), a constant value can be added to $P(\event | \metamodel, \x)$.
$\metamodel$ can also be a regression model predicting some measure correlating with $J(\event)$ (e.g., the minimum distance between two potentially colliding objects). Here, models like Gaussian processes (GP) naturally ensure $P(\event | \metamodel, \x)>0$ since the cumulative density function used for calculating $P(\event | \metamodel, \x)$ is strictly greater than $0$.

To sample from the ``non-parametric multidimensional pseudo-PDF'', \cite[p.~11]{dubourg_metamodel-based_2013} uses slice sampling.
For TIS, we use a different procedure: Since $q_{\metamodel}(\x) \propto P(\event | \metamodel, \x)\ p(\x)$, we sample from $q_{\metamodel}(\x)$ by first sampling from $p(\x)$ and accepting the sample with $P(\event | \metamodel, \x)$.
No likelihoods have to be calculated. Thus, with good availability of data, $p(\x)$ can simply be a database of scenario parameterizations (e.g., from field studies) whose content does not have to be approximated by density estimation (a source of bias).

For risk estimation, we insert \cref{eq:q_mm} into \cref{eq:is_mean} resulting in \cref{eq:mm_mean}.
For a detailed discussion of the variance of this estimator, we refer to~\cite{dubourg_metamodel-based_2013}. Practically, a high correlation of $J$ and $P$ leads to a low variance, which would vanish to $0$ if $J(\event | \model, \x_i) / P(\event | \metamodel, \x_i) = \text{constant}\ \forall\ i \in \mathbb{N}^{\leq N_\ell}$.
\begin{equation} \label{eq:mm_mean}
	\ell_{\event \model} = \int P(\event | \metamodel, \x)\ p(\x)\ d\x\ \frac{1}{N_\ell} \sum_{i=1}^{N_\ell} \frac{J(\event | \model, \x_i)}{P(\event | \metamodel, \x_i)}
\end{equation}

\begin{figure}[!t]
	\centering
	\includegraphics[width=3.4in]{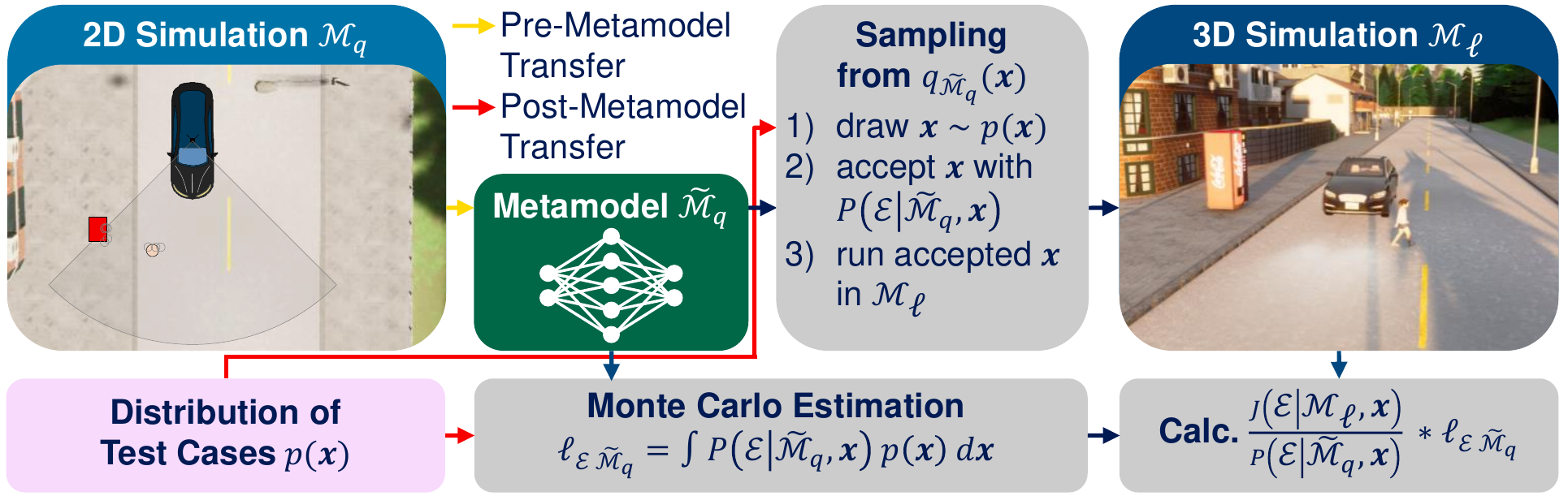}
	\caption{Depiction of the process for transfer importance sampling, explained in detail in \cref{subsec:performant_setup}. If $\model_\deriveq$ and $\model_\estimateell$ use different parameters, there must be a transfer between the models, which is explained in \cref{subsec:hotis_hetis}.} 	\label{fig:process}
\end{figure}

\subsection{Risk Estimation Linking Two Test Setups}\label{subsec:performant_setup}
TIS builds on the idea of metamodel-based IS and involves the following steps illustrated in \cref{fig:process}:
Firstly, a metamodel $\metamodel_\deriveq$ is built based on $N_q$ test runs in $\model_\deriveq$ (e.g., a scalable 2D simulation) and used for an MC estimation of $\ell_{\event \metamodel_\deriveq}$.

Secondly, $N_\ell$ test parameterizations $\x_i\ \forall\ i \in \mathbb{N}^{\leq N_\ell}$ are drawn from the proposal density in \cref{eq:q_tis} and carried out in $\model_\estimateell$ (e.g., a detailed but costly 3D simulation).
\begin{equation} \label{eq:q_tis}
	q_{\metamodel_\deriveq}(\x) = \frac{P(\event | \metamodel_\deriveq, \x)\ p(\x)}{\int P(\event | \metamodel_\deriveq, \x)\ p(\x)\ d\x} = \frac{P(\event | \metamodel_\deriveq, \x)\ p(\x)}{\ell_{\event \metamodel_\deriveq}}
\end{equation}
Analogous to \cref{subsec:mmis}, $P(\event | \metamodel_\deriveq, \x)>0$ can be easily ensured.
Yet, already $P(\event | \metamodel_\deriveq, \x)$ close to $0$ can cause an IS estimate to initially converge to some value and later increase significantly (e.g. if there is a cluster of accidents in $\model_\estimateell$ not occurring in $\model_\deriveq$).
To hedge against this case, $P(\event | \metamodel_\deriveq, \x)$ can be skewed by increasing small values. However, this increases the frequency of non-critical test runs, resulting in a tradeoff between reliability and performance.
To assess this tradeoff, we refrain from skewing in our experiments.

Finally, \cref{eq:tis_mean} links $\metamodel_\deriveq$ and $\model_\estimateell$ for the IS risk estimate.
\begin{equation} \label{eq:tis_mean}
	\ell_{\event \model_\estimateell} = \int P(\event | \metamodel_\deriveq, \x)\ p(\x)\ d\x\ \frac{1}{N_\ell} \sum_{i=1}^{N_\ell} \frac{J(\event | \model_\estimateell, \x_i)}{P(\event | \metamodel_\deriveq, \x_i)}
\end{equation}

\subsection{Homogeneous and Heterogeneous Transfer}\label{subsec:hotis_hetis}
Using two test setups, there is an important consideration:
In the field of transfer learning~\cite{day_survey_2017}, homogeneous TL describes applications where different data sources share the same representation, i.e., a scenario is described using the same parameters in $\model_\estimateell$ and $\model_\deriveq$.
In heterogeneous TL, the representations differ.
For example, weather conditions in $\model_\estimateell$ might be represented by sensor noise in $\model_\deriveq$.
As shown in \cref{fig:process}, we can handle heterogeneous cases in two ways, which we explain in detail in \cref{subsec:uc_1} and \cref{subsec:uc_2}:

\textit{1) Pre-Metamodel Transfer} is applicable in a verification \& validation phase, where the parameters $\x_\estimateell$ to be varied in $\model_\estimateell$ are known. Here, before building $\metamodel_\deriveq$, $\model_\deriveq$ can be extended so that its test runs are parameterized by $\x_\estimateell$.

\textit{2) Post-Metamodel Transfer} allows using test results from a concept phase, where the parameters $\x_\estimateell$ to be varied in $\model_\estimateell$ might be unknown. Here, $\metamodel_\deriveq$ is built based on parameters $\x_\deriveq$.
To let $\metamodel_\deriveq$ predict test runs $\x_\estimateell$, these are first passed through a transfer function $\transfer$, i.e., $\x_\deriveq = \transfer(\x_\estimateell)$.

\begin{figure}[!t]
	\centering
	\includegraphics[width=2.0in]{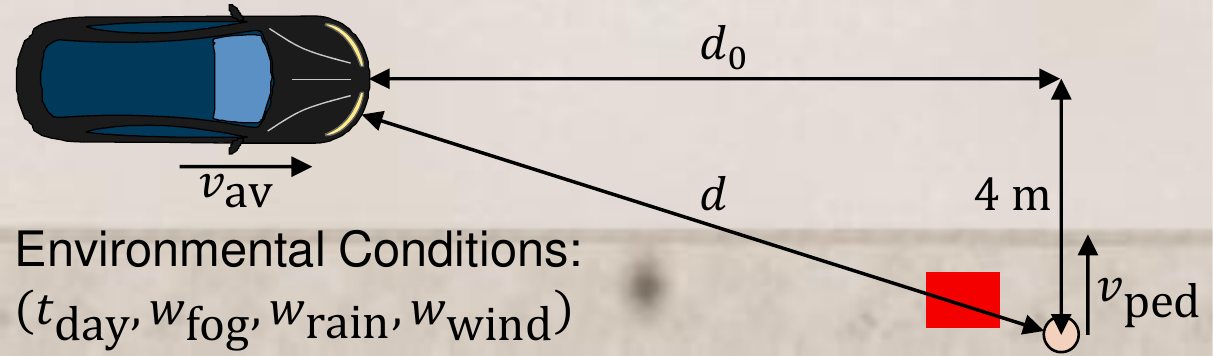}
	\caption{Depiction of the jaywalking scenario. The AV starts from standstill and accelerates to $v_\mathrm{av}$. As soon as the AV arrives at a triggering point (at which it has reached $v_\mathrm{av}$), an initially occluded pedestrian at a longitudinal distance of $d_0$ starts moving towards the road with a velocity of $v_\mathrm{ped}$. All variables' units, ranges, and the environmental conditions are described in \cref{subsec:jaywalking}. The distance $d$ is used for criticality assessment (see \cref{eq:d_min_extended}).} 		\label{fig:scenario}
\end{figure}

\section{Experiments}
The goal of the experiments is to demonstrate the proposed method while exploring its limits of applicability.
Below, we describe the components of our experimental setup.

\subsection{The Jaywalking Scenario}\label{subsec:jaywalking}
We investigate the jaywalking scenario in \cref{fig:scenario}.
To parameterize the environmental conditions, we fused pedestrian counting data~\cite{sentoff_vermont_2017} with weather data. As a result, we have hourly pedestrian counts and weather conditions over a period of 8.5 years.
Assuming that the probability of jaywalking is proportional to the pedestrian volume, the conditions drawn from the dataset mostly correspond to daylight and clear weather.
Since the pedestrian counts are aggregated over full hours, we choose a specific time of day uniformly distributed within the bins.
To apply the weather conditions in simulation, we norm $w_\mathrm{fog}, w_\mathrm{wind}$, and $w_\mathrm{rain}$ to $[0, 1]$.
The tire-road friction is $\mu_\mathrm{fric}(w_\mathrm{rain}) = 0.5 + 0.4\ e^{-20\ w_\mathrm{rain}}$, resulting in $\mu_\mathrm{fric}(0) = 0.9$ for dry conditions and converging to $\mu_\mathrm{fric}(1) = 0.5$ for the most intense rain over the data collection period of 8.5 years.

Regardless of the environmental conditions, the AV has a defensive target velocity of $v_\mathrm{av} \sim \mathcal{N}(6\ \mathrm{m/s}, 0.2\ \mathrm{m/s})$.
A model describing at which traffic gaps pedestrians jaywalk~\cite{wang_study_2010} is used to determine $d_0$; we use the parameters for young pedestrians.
Following~\cite{movahhed_effect_2020}, the pedestrian's velocity $v_\mathrm{ped}$ depends on $w_\mathrm{rain}$; the mean and standard deviation for female pedestrians are interpolated linearly based on $w_\mathrm{rain}$.

Although we have enough data to assess the scenario without density estimation (see \cref{subsec:mmis}), we consider the usual case of limited data. Hence, we drew 10,000 samples from our dataset and optimized a multivariate kernel density estimate (KDE) using maximum likelihood cross-validation. Sampling from the KDE, we discard both invalid test runs (there may be slightly negative values, e.g., for $w_\mathrm{fog}$) and test runs with $d_0 > 50\ \mathrm{m}$, since with the given distribution, collisions above that distance are almost impossible.

The criticality of test runs is assessed using \cref{eq:d_min_extended}, where $d_\mathrm{min}$ is the minimum of all frames' $d$. Thereby, the AV is approximated by a rectangle and the pedestrian by a circle. Since for all collisions, $d_\mathrm{min} = 0$, we treat this case by calculating the theoretically remaining braking distance after the collision. Accordingly, $v_\mathrm{av\ col}$ is the AV's velocity at the initial collision and $g$ is Earth's gravity. As a result, a smooth transition between non-collisions and collisions is achieved.
\begin{equation} \label{eq:d_min_extended}
	d^*_\mathrm{min}=\begin{cases}
		d_\mathrm{min},                                   & \text{if $d_\mathrm{min} > 0$}. \\
		- v^2_\mathrm{av\ col}/(2\ g\ \mu_\mathrm{fric}), & \text{otherwise}.
	\end{cases}
\end{equation}

\subsection{3D Software in the Loop for the Actual Risk Estimation}
In our demonstration, $\model_\estimateell$ involves the modular AV stack Pylot~\cite{gog_pylot_2021} and the 3D simulation tool CARLA~\cite{dosovitskiy_carla_2017}. In this setup, we use Pylot's perception, planning, and control components. We hence denote the setup as 3D SiL (Software in the Loop).
In CARLA, $t_\mathrm{day}$ influences the angle of the sun and we directly apply $(w_\mathrm{fog}, w_\mathrm{wind}, w_\mathrm{rain}, \mu_\mathrm{fric})$.
Using an NVIDIA GeForce RTX 2080 Ti, one test run takes about 90 seconds.
Test results and videos are available online\footnote{\url{https://github.com/wnklmx/DSIOD/tree/main/data/202202_Jaywalking}}.

To use AIS (metamodel-based IS) as a baseline, we metamodel the AV's behavior in the 3D SiL.
We use a GP model with the hyperparameters stated in our own prior work~\cite{winkelmann_probabilistic_2021}, where GPs achieved reliable uncertainties.
The GP is trained with $N_q = 200$ test runs from the Sobol sequence~\cite{sobol_distribution_1967}, distributed over a parameter space encompassing the largest part of the KDE's samples; training takes about $5\ \mathrm{s}$.
Based on \cref{eq:d_min_extended}, $P(\event | \metamodel, \x)$ is chosen as $P(d^*_\mathrm{min} < 0 | \metamodel, \x)$.
The metamodels are built the same way for the other test setups.

\subsection{Transfer Based on 2D Software in the Loop}\label{subsec:uc_1}
To examine pre-metamodel transfer, $\model_\deriveq$ involves the 2D simulation tool IAV Scene Suite and the planning and control components of Pylot (compare \cref{fig:test_setups}). We supply Pylot with an OpenDrive map, a pose, and an object list containing  object types and (perceived) positions. We denote this setup as 2D SiL and the overall use-case as 2D SiL TIS.

To create a perception model, we used CARLA to render images with a pedestrian and varying weather conditions and evaluated \textit{if} and \textit{where} Pylot detects the pedestrian. An extra-trees~\cite{geurts_extremely_2006} classification model predicts if the pedestrian is detected based on the relative position of the pedestrian with respect to the AV and $(t_\mathrm{day}, w_\mathrm{fog}, w_\mathrm{rain})$. If the pedestrian is detected, an extra-trees regression model uses the same features to determine the perceived position.
Since extra-trees' predictions are discrete distributions, we choose a random detection and regression model per frame, increasing efficiency and allowing to handle non-Gaussian sensor noise.

We also metamodeled CARLA's dynamics model: We used the velocity, acceleration, throttle, and brake values of 4 consecutive time steps to build an extra-trees regression model predicting the acceleration in the next time step. Here, we use the mean of the ensemble's predictions.
The previous steps are obsolete if suitable component models (data- or knowledge-based) are available (e.g., in large AV projects).

\begin{figure}[!t]
	\centering
	\includegraphics[width=3.2in]{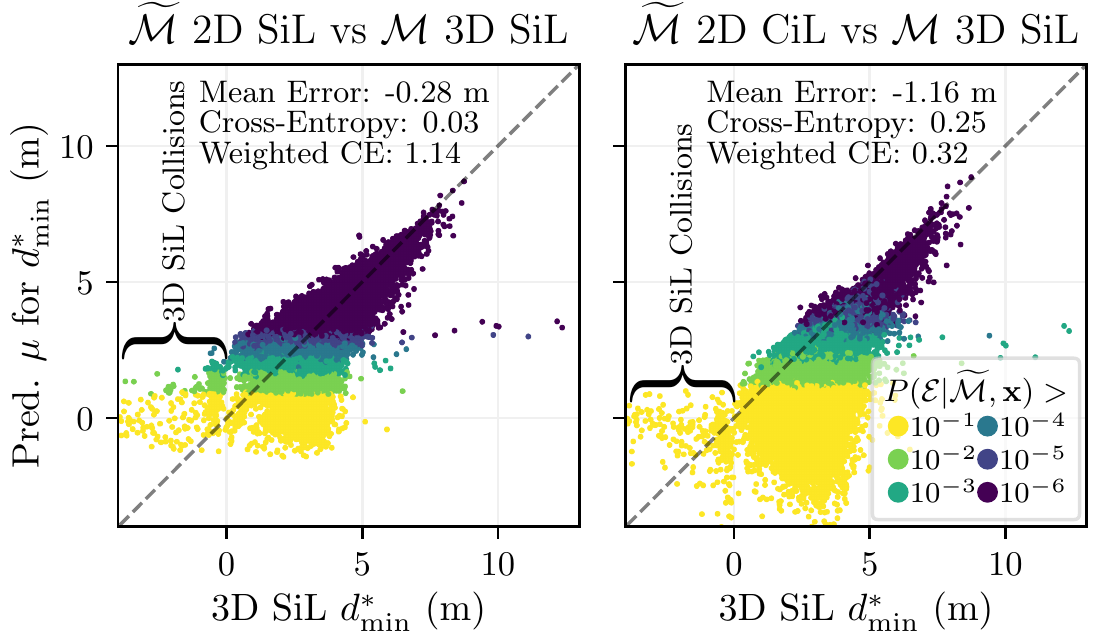}
	\caption{Comparison of the 3D SiL's outcomes to the metamodels' predictions. The samples are drawn from $p(\x)$ and colored based on the predicted probability of a collision. The 2D SiL's metamodel reflects the trend of criticality quite well. The 2D CiL's metamodel predicts more critical outcomes, which, however, leads to a better weighted CE (Cross-Entropy).}
	\label{fig:correlation}
\end{figure}

\subsection{Transfer Based on 2D Concept in the Loop}\label{subsec:uc_2}
To examine post-metamodel transfer, $\model_\deriveq$ again involves IAV Scene Suite, but none of Pylot's components.
The perception model has a frame-wise detection probability $P_\mathrm{detect} \in [0.4, 1]$ decreasing with the pedestrian’s distance~\cite{yang_obstacles_2014}, a reaction time of $0.4\ \mathrm{s}$, and the pedestrian's true relative distance is multiplied with $\mathcal{N}(1, \sigma_\mathrm{noise})$, where $\sigma_\mathrm{noise} \in [0, 0.05]$.
A simple planning module is used: The AV's target speed is set to $0$ whenever a pedestrian is detected in or moving towards the AV's lane. The controller is a PID controller.
The dynamics model limits the AV's deceleration so that it linearly approaches the minimum of $-g\ \mu_\mathrm{fric}$ over $0.2\ \mathrm{s}$, where $\mu_\mathrm{fric} \in [0.5, 1]$.
We denote this setup as 2D CiL (Concept in the Loop) and the use-case as 2D CiL TIS.

According to above models, the metamodel is built using $\x_\deriveq = (d_0, v_\mathrm{av}, v_\mathrm{ped}, P_{\mathrm{detect}}, \sigma_\mathrm{noise}, \mu_\mathrm{fric})$; the parameterizations are distributed evenly within the stated ranges. We simply guess the transfer function $\transfer$: $(d_0, v_\mathrm{av}, v_\mathrm{ped})$ is not modified,
$P_{\mathrm{detect}} = 1 - 0.4\ w_\mathrm{fog} - 0.4\ w_\mathrm{rain} - 0.2\ \frac{|t_\mathrm{day} - 12\ \mathrm{h}|}{12\ \mathrm{h}}$,
$\sigma_\mathrm{noise}=0.03$, and
$\mu_\mathrm{fric} = 0.5 + 0.4\ e^{-20\ w_\mathrm{rain}}$.
\begin{figure*}[!t]
	\centering
	\includegraphics[width=7in]{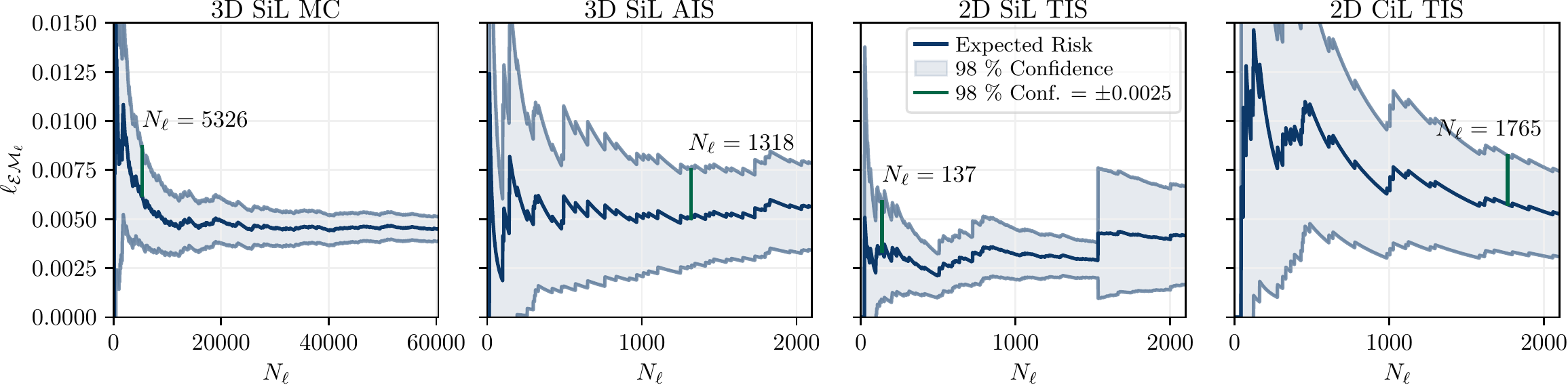}
	\caption{Convergence of the risk estimates. Using the 3D SiL only, MC (Monte Carlo) sampling converges the slowest and AIS (adaptive importance sampling) converges 4 times as fast. 2D SiL TIS (transfer importance sampling) converges even faster than AIS – until at 1533 test runs, the expected risk and variance suddenly increase; this indicates an unexpected collision (see \cref{subsec:performant_setup}). 2D CiL TIS converges stably and slightly slower than AIS.}
	\label{fig:convergence}
\end{figure*}
\section{Results}
Below, we evaluate the quality of the metamodels as well as the convergence and efficiency of the risk estimation.

\subsection{Predictive Performance of the Metamodels}\label{subsec:ppmm}
\cref{fig:correlation} compares the metamodels' predictions to the outcomes of the 3D SiL.
The mean error indicates that the 2D SiL metamodel's predictions are more accurate; the CE, assessing the metamodels' classification of the 3D SiL's outcomes as collision or collision-free, is lower (better) as well.
However, the bias of the 2D CiL's metamodel towards more critical outcomes leads to a better weighted CE (which weights the correct prediction of the 3D SiL's collisions higher than the correct predictions of collision-free test runs).

From these results, we expect that with the 2D CiL's metamodel, a higher number of collision-free test parameterizations will be sampled. With the 2D SiL's metamodel, the fraction of collision-free samples will be lower but there might be surprising collisions (see \cref{subsec:performant_setup}).

\subsection{Convergence of the Risk Estimates}\label{subsec:convergence}
As explained in \cref{subsec:mmis}, the metamodels are used for an MC estimation of the expected risk. Here, the 2D SiL's metamodel leads to $\ell_{\event \metamodel_\deriveq} \approx 2.4~\%$ and the 2D CiL's metamodel leads to $\ell_{\event \metamodel_\deriveq} \approx 26.4~\%$.
Despite these different values, \cref{fig:convergence} shows that for the 3D SiL, all approaches' estimates converge to an expected risk of $\ell_{\event \model_\estimateell} \approx 0.45~\%$.

On closer inspection, \cref{fig:convergence} reveals what was expected in \cref{subsec:ppmm}. 2D CiL TIS converges slowly (since there are many collision-free test runs). 2D SiL TIS converges faster – until an unexpected collision leads to a large quotient of $J(\event | \model_\estimateell, \x_i)/P(\event | \metamodel_\deriveq, \x_i)$ in \cref{eq:tis_mean} and a sudden increase of the expected risk and variance.
The 3D SiL's results can explain this: about $70~\%$ of the collisions occur for $t_\mathrm{day} \in [19~\mathrm{h}, 21~\mathrm{h}]$, where the sun is low and blinds the AV's camera.
Such collisions, which depend to a large extent on a phenomenon that mainly occurs in the 3D SiL, are difficult to predict based on the 2D SiL and 2D CiL.
Still, 2D SiL TIS leads to frequent collisions and converges the fastest.

At its minimum at 495 test runs, the upper confidence interval of the 2D SiL TIS estimate is about $30~\%$ below the true risk. Clearly, such a bias is undesirable and hedging against small $P(\event | \metamodel_\deriveq, \x)$ can be crucial to prevent early convergence (see \cref{subsec:performant_setup}).
Nevertheless, the alternative of evaluating risks only in cost-effective test setups would often be much worse: In our case, using only the 2D SiL or 2D CiL would result in a bias of factor 5 or 59, respectively.

\subsection{Comparison of the Approaches' Efficiency}
To compare the approaches' efficiency, we aim for
a $1~\%$ chance that the true $\ell_{\event \model_\estimateell}$ is more than $50~\%$ larger than the estimated $\ell_{\event \model_\estimateell}$. \cref{fig:convergence} shows the number $N_\ell$ of required 3D SiL test runs, but the $N_q$ test runs used to derive $q(\x)$ must be considered as well.
A test run in the 3D SiL, 2D SiL, and 2D CiL takes $90~\mathrm{s}$, $75~\mathrm{s}$, and $27~\mathrm{s}$, respectively.
Using an Amazon Web Services (AWS) g4dn.2xlarge instance (0.752~\$/h, with GPU) for the 3D SiL and an AWS t4g.2xlarge instance (0.2688~\$/h, without GPU) for the 2D SiL and 2D CiL, the costs correspond to \cref{tab:cost} (prices\footnote{\url{https://aws.amazon.com/ec2/pricing/on-demand/}} from February 2022).
2D SiL TIS results in the lowest costs, AIS is 7.8 times, and 2D CiL TIS is 9.2 times as expensive.
Such a comparison depends on the scenario, its risk level, the similarity of the results in the two test setups, and the test setups' difference in cost. For other test setups, pure time savings may also be more important. Hence, the potential savings increase for homogeneous transfer or if, e.g., a parallelized 3D SiL is linked to costly and time-consuming proving ground tests.

Overall, it shows that using only the 2D SiL or 2D CiL would lead to a biased estimate and that the AIS estimation using only the 3D SiL is not the cheapest option.
Hence, 2D SiL TIS (linking the 2D SiL and 3D SiL) performs best.

\begin{table}[!h]
		\renewcommand{\arraystretch}{1.18}
	\setlength{\tabcolsep}{4pt}
	\begin{center}
		\caption{Efficiency Comparison ($N_q = 200$)}
		\label{tab:cost}
		\begin{tabular}{|l|r|r|r|r|r|}
			\hline
			\textbf{Approach} & \textbf{Cost} $N_q$ & $N_\ell$ & \textbf{Cost} $N_\ell$ & \textbf{Total Cost} & \textbf{Factor} \\
			\hline
			3D SiL MC       & 0 \$                & 5326     & 100.13 \$              & 100.13 \$           & 27.4            \\
			\hline
			3D SiL AIS      & 3.76 \$             & 1318     & 24.78 \$               & 28.54 \$            & 7.8             \\
			\hline
			2D SiL TIS      & 1.08 \$             & 137      & 2.57 \$                & 3.65 \$             & 1               \\
			\hline
			2D CiL TIS      & 0.40 \$             & 1765     & 33.19 \$               & 33.59 \$            & 9.2             \\
			\hline
		\end{tabular}
	\end{center}
\end{table}

\section{Conclusion}
This paper presented transfer importance sampling, a risk estimation method linking different test setups.
While existing methods from the field of (adaptive) importance sampling reduce a risk estimate's variance, they sometimes require so many test runs that scalable, purely virtual test setups must be used – which may lead to biased risk estimates.
Transfer importance sampling uses a scalable test setup for an exploratory analysis and can then estimate risks with a small number of test runs in a more trustworthy, possibly costly test setup. This way, bias and variance can be reduced.

In our experiments, we examined two use-cases where the risk estimation (in a 3D simulation) involved influences that could not simply be modeled in the preceding exploratory analysis (using a 2D simulation).
Transfer importance sampling outperforms adaptive importance sampling if the used test setups show similar behavior. Conversely, it performs worse if the used test setups differ too much in their behavior.
Hence, we aim to extend our approach to incorporate the test runs of the actual risk estimation into the metamodel. This could increase efficiency and further avoid early convergence of a risk estimate due to the results of the cost-effective (potentially abstract) test setup putting the risk estimation process on the wrong track (see \cref{subsec:convergence}).

Applying transfer importance sampling is particularly reasonable for scenarios with higher risk levels, where biased risk estimates can have fatal consequences.
Thereby, it is usually not obvious which test setup leads to low bias.
Not without reason, ISO 21448~\cite{iso_central_secretary_road_2019} requires that risks must be continuously monitored during an AV's operation phase.
Yet, a sheer comparison of the results of different test setups can be valuable in itself.
The ultimate challenge is the transition from an AV's verification \& validation phase to its operations phase.
This transition will be easier if transitions between 2D simulations, 3D simulations, and real tests have been mastered before.
The concepts presented are valuable for making such transitions.
Here, besides standards for scenarios (e.g., OpenDRIVE, OpenSCENARIO, OpenODD), standards for AV component models could reduce the overhead of investigating scenarios in multiple test setups.

Conclusively, it can be said that transfer importance sampling can help to estimate risks posed by AV more accurately and efficiently.
With our method, we take an important step in uniting test setups and results along the development process to support the creation of evidence for an AV's release.

\bibliographystyle{IEEEtran}
\bibliography{IEEEabrv,./refs/refs}

\begin{thebibliography}{10}
\providecommand{\url}[1]{#1}
\csname url@samestyle\endcsname
\providecommand{\newblock}{\relax}
\providecommand{\bibinfo}[2]{#2}
\providecommand{\BIBentrySTDinterwordspacing}{\spaceskip=0pt\relax}
\providecommand{\BIBentryALTinterwordstretchfactor}{4}
\providecommand{\BIBentryALTinterwordspacing}{\spaceskip=\fontdimen2\font plus
\BIBentryALTinterwordstretchfactor\fontdimen3\font minus
  \fontdimen4\font\relax}
\providecommand{\BIBforeignlanguage}[2]{{%
\expandafter\ifx\csname l@#1\endcsname\relax
\typeout{** WARNING: IEEEtran.bst: No hyphenation pattern has been}%
\typeout{** loaded for the language `#1'. Using the pattern for}%
\typeout{** the default language instead.}%
\else
\language=\csname l@#1\endcsname
\fi
#2}}
\providecommand{\BIBdecl}{\relax}
\BIBdecl

\bibitem{iso_central_secretary_road_2019}
{ISO Central Secretary}, ``\BIBforeignlanguage{en}{Road vehicles — {Safety}
  of the intended functionality},'' International Organization for
  Standardization, Geneva, CH, Standard ISO/PAS 21448:2019, 2019.

\bibitem{kalra_driving_2016}
N.~Kalra and S.~Paddock, ``Driving to safety: {How} many miles of driving would
  it take to demonstrate autonomous vehicle reliability?'' \emph{Transportation
  Research Part A: Policy and Practice}, vol.~94, pp. 182--193, Dec. 2016.

\bibitem{riedmaier_survey_2020}
S.~Riedmaier, T.~Ponn, D.~Ludwig, B.~Schick, and F.~Diermeyer, ``Survey on
  {Scenario}-{Based} {Safety} {Assessment} of {Automated} {Vehicles},''
  \emph{IEEE Access}, vol.~8, pp. 87\,456--87\,477, 2020.

\bibitem{huang_autonomous_2016}
W.~Huang, K.~Wang, Y.~Lv, and F.~Zhu, ``Autonomous vehicles testing methods
  review,'' in \emph{2016 {IEEE} 19th {International} {Conference} on
  {Intelligent} {Transportation} {Systems} ({ITSC})}.\hskip 1em plus 0.5em
  minus 0.4em\relax IEEE, Nov. 2016, pp. 163--168, iSSN: 2153-0017.

\bibitem{steimle_toward_2022}
M.~Steimle, N.~Weber, and M.~Maurer, ``Toward {Generating} {Sufficiently}
  {Valid} {Test} {Case} {Results}: {A} {Method} for {Systematically}
  {Assigning} {Test} {Cases} to {Test} {Bench} {Configurations} in a
  {Scenario}-{Based} {Test} {Approach} for {Automated} {Vehicles},'' \emph{IEEE
  Access}, vol.~10, pp. 6260--6285, 2022.

\bibitem{huang_synthesis_2018}
Z.~Huang, M.~Arief, H.~Lam, and D.~Zhao, ``Synthesis of {Different}
  {Autonomous} {Vehicles} {Test} {Approaches},'' in \emph{2018 21st
  {International} {Conference} on {Intelligent} {Transportation} {Systems}
  ({ITSC})}.\hskip 1em plus 0.5em minus 0.4em\relax IEEE, Nov. 2018, pp.
  2000--2005, iSSN: 2153-0017.

\bibitem{yang_development_2010}
H.-H. Yang and H.~Peng, ``Development and evaluation of collision
  warning/collision avoidance algorithms using an errable driver model,''
  \emph{Vehicle System Dynamics}, vol.~48, no. sup1, pp. 525--535, Dec. 2010.

\bibitem{woodrooffe_evaluation_2014}
J.~Woodrooffe and D.~Blower, ``\BIBforeignlanguage{en}{Evaluation of forward
  collision mitigation braking safety performance for commercial vehicles},''
  in \emph{\BIBforeignlanguage{en}{13th {International} {Symposium} on {Heavy}
  {Vehicle} {Transportation} {Technology}}}, Argentinia, 2014.

\bibitem{gietelink_probabilistic_2005}
O.~Gietelink, B.~De~Schutter, and M.~Verhaegen,
  ``\BIBforeignlanguage{en}{Probabilistic {Validation} of {Advanced} {Driver}
  {Assistance} {Systems}},'' \emph{\BIBforeignlanguage{en}{IFAC Proceedings
  Volumes}}, vol.~38, no.~1, pp. 97--102, Jan. 2005.

\bibitem{zhao_accelerated_2016}
D.~Zhao, H.~Peng, S.~Bao, K.~Nobukawa, D.~LeBlanc, and C.~Pan, ``Accelerated
  evaluation of automated vehicles using extracted naturalistic driving data,''
  in \emph{The {Dynamics} of {Vehicles} on {Roads} and {Tracks}}.\hskip 1em
  plus 0.5em minus 0.4em\relax CRC Press, Apr. 2016, pp. 287--296.

\bibitem{akagi_risk-index_2019}
Y.~Akagi, R.~Kato, S.~Kitajima, J.~Antona-Makoshi, and N.~Uchida, ``A
  {Risk}-index based {Sampling} {Method} to {Generate} {Scenarios} for the
  {Evaluation} of {Automated} {Driving} {Vehicle} {Safety},'' in \emph{2019
  {IEEE} {Intelligent} {Transportation} {Systems} {Conference} ({ITSC})}.\hskip
  1em plus 0.5em minus 0.4em\relax Auckland, New Zealand: IEEE Press, Oct.
  2019, pp. 667--672.

\bibitem{wang_combining_2020}
X.~Wang, H.~Peng, and D.~Zhao, ``Combining {Reachability} {Analysis} and
  {Importance} {Sampling} for {Accelerated} {Evaluation} of {Highway}
  {Automated} {Vehicles} at {Pedestrian} {Crossing},'' \emph{ASME Letters in
  Dynamic Systems and Control}, vol.~1, no.~1, Mar. 2020.

\bibitem{rubinstein_cross-entropy_1999}
R.~Rubinstein, ``\BIBforeignlanguage{en}{The {Cross}-{Entropy} {Method} for
  {Combinatorial} and {Continuous} {Optimization}},''
  \emph{\BIBforeignlanguage{en}{Methodology And Computing In Applied
  Probability}}, vol.~1, no.~2, pp. 127--190, Sep. 1999.

\bibitem{zhao_accelerated_2017}
D.~Zhao, H.~Lam, H.~Peng, S.~Bao, D.~J. LeBlanc, K.~Nobukawa, and C.~S. Pan,
  ``Accelerated {Evaluation} of {Automated} {Vehicles} {Safety} in
  {Lane}-{Change} {Scenarios} {Based} on {Importance} {Sampling}
  {Techniques},'' \emph{IEEE Transactions on Intelligent Transportation
  Systems}, vol.~18, no.~3, pp. 595--607, Mar. 2017.

\bibitem{okelly_scalable_2018}
M.~O'Kelly, A.~Sinha, H.~Namkoong, R.~Tedrake, and J.~C. Duchi, ``Scalable
  {End}-to-{End} {Autonomous} {Vehicle} {Testing} via {Rare}-event
  {Simulation},'' in \emph{Advances in {Neural} {Information} {Processing}
  {Systems}}, vol.~31.\hskip 1em plus 0.5em minus 0.4em\relax Curran
  Associates, Inc., 2018.

\bibitem{huang_accelerated_2018}
Z.~Huang, H.~Lam, D.~J. LeBlanc, and D.~Zhao, ``Accelerated {Evaluation} of
  {Automated} {Vehicles} {Using} {Piecewise} {Mixture} {Models},'' \emph{IEEE
  Transactions on Intelligent Transportation Systems}, vol.~19, no.~9, pp.
  2845--2855, Sep. 2018.

\bibitem{jesenski_creation_2021}
S.~Jesenski, N.~Tiemann, W.~Branz, and J.~M. Zöllner, ``Creation of {Critical}
  {Traffic} {Scenes} for {Usage} with {Importance} {Sampling},'' in \emph{2021
  {IEEE} {International} {Intelligent} {Transportation} {Systems} {Conference}
  ({ITSC})}.\hskip 1em plus 0.5em minus 0.4em\relax IEEE, Sep. 2021, pp.
  3162--3169.

\bibitem{gietelink_adaptive_2006}
O.~Gietelink, B.~De~Schutter, and M.~Verhaegen, ``Adaptive importance sampling
  for probabilistic validation of advanced driver assistance systems,'' in
  \emph{2006 {American} {Control} {Conference}}, Jun. 2006, pp. 4002--4007,
  iSSN: 2378-5861.

\bibitem{de_gelder_assessment_2017}
E.~de~Gelder and J.-P. Paardekooper, ``Assessment of {Automated} {Driving}
  {Systems} {Using} {Real}-{Life} {Scenarios},'' in \emph{2017 {IEEE}
  {Intelligent} {Vehicles} {Symposium} ({IV})}.\hskip 1em plus 0.5em minus
  0.4em\relax IEEE, Jun. 2017, pp. 589--594.

\bibitem{norden_efficient_2020}
J.~Norden, M.~O'Kelly, and A.~Sinha, ``Efficient {Black}-box {Assessment} of
  {Autonomous} {Vehicle} {Safety},'' \emph{arXiv:1912.03618 [cs, stat]}, Jun.
  2020.

\bibitem{dubourg_metamodel-based_2013}
V.~Dubourg, B.~Sudret, and F.~Deheeger,
  ``\BIBforeignlanguage{en}{Metamodel-based importance sampling for structural
  reliability analysis},'' \emph{\BIBforeignlanguage{en}{Probabilistic
  Engineering Mechanics}}, vol.~33, pp. 47--57, Jul. 2013.

\bibitem{day_survey_2017}
O.~Day and T.~M. Khoshgoftaar, ``A survey on heterogeneous transfer learning,''
  \emph{Journal of Big Data}, vol.~4, no.~1, p.~29, Sep. 2017.

\bibitem{sentoff_vermont_2017}
K.~Sentoff and J.~L. Sullivan, ``\BIBforeignlanguage{English}{Vermont {Bicycle}
  and {Pedestrian} {Counting} {Program}},'' University of Vermont
  Transportation Research Center, Montpelier, VT United States, Publication TRC
  Report 17-006, Nov. 2017.

\bibitem{wang_study_2010}
T.~Wang, J.~Wu, P.~Zheng, and M.~McDonald, ``Study of pedestrians' gap
  acceptance behavior when they jaywalk outside crossing facilities,'' in
  \emph{13th {International} {IEEE} {Conference} on {Intelligent}
  {Transportation} {Systems}}.\hskip 1em plus 0.5em minus 0.4em\relax IEEE,
  Sep. 2010, pp. 1295--1300, iSSN: 2153-0017.

\bibitem{movahhed_effect_2020}
M.~B. Movahhed, J.~Ayoubinejad, F.~N. Asl, and M.~Feizbahr,
  ``\BIBforeignlanguage{en}{The {Effect} of {Rain} on {Pedestrians} {Crossing}
  {Speed}},'' \emph{\BIBforeignlanguage{en}{Computational Research Progress in
  Applied Science \& Engineering}}, vol.~06, no.~03, pp. 186--190, 2020.

\bibitem{gog_pylot_2021}
I.~Gog, S.~Kalra, P.~Schafhalter, M.~A. Wright, J.~E. Gonzalez, and I.~Stoica,
  ``Pylot: {A} {Modular} {Platform} for {Exploring} {Latency}-{Accuracy}
  {Tradeoffs} in {Autonomous} {Vehicles},'' in \emph{2021 {IEEE}
  {International} {Conference} on {Robotics} and {Automation} ({ICRA})}.\hskip
  1em plus 0.5em minus 0.4em\relax IEEE, May 2021, pp. 8806--8813, iSSN:
  2577-087X.

\bibitem{dosovitskiy_carla_2017}
A.~Dosovitskiy, G.~Ros, F.~Codevilla, A.~Lopez, and V.~Koltun, ``{CARLA}: {An}
  {Open} {Urban} {Driving} {Simulator},'' in \emph{Proceedings of the 1st
  {Annual} {Conference} on {Robot} {Learning}}, ser. Proceedings of {Machine}
  {Learning} {Research}, S.~Levine, V.~Vanhoucke, and K.~Goldberg, Eds.,
  vol.~78.\hskip 1em plus 0.5em minus 0.4em\relax PMLR, Nov. 2017, pp. 1--16.

\bibitem{winkelmann_probabilistic_2021}
M.~Winkelmann, M.~Kohlhoff, H.~H. Tadjine, and S.~Müller, ``Probabilistic
  {Metamodels} for an {Efficient} {Characterization} of {Complex} {Driving}
  {Scenarios},'' \emph{arXiv:2110.02892 [cs, eess]}, Oct. 2021.

\bibitem{sobol_distribution_1967}
I.~M. Sobol', ``\BIBforeignlanguage{en}{On the distribution of points in a cube
  and the approximate evaluation of integrals},''
  \emph{\BIBforeignlanguage{en}{USSR Computational Mathematics and Mathematical
  Physics}}, vol.~7, no.~4, pp. 86--112, Jan. 1967.

\bibitem{geurts_extremely_2006}
P.~Geurts, D.~Ernst, and L.~Wehenkel, ``\BIBforeignlanguage{en}{Extremely
  randomized trees},'' \emph{\BIBforeignlanguage{en}{Machine Learning}},
  vol.~63, no.~1, pp. 3--42, Apr. 2006.

\bibitem{yang_obstacles_2014}
Y.~Yang, R.~Mingwu, and Y.~Jingyu, ``\BIBforeignlanguage{en}{Obstacles and
  {Pedestrian} {Detection} on a {Moving} {Vehicle}},''
  \emph{\BIBforeignlanguage{en}{International Journal of Advanced Robotic
  Systems}}, vol.~11, no.~4, p.~53, Apr. 2014, publisher: SAGE Publications.

\end{thebibliography}

\end{document}